\documentclass[letterpaper]{article} 
\usepackage{aaai25}  
\usepackage{times}  
\usepackage{helvet}  
\usepackage{courier}  
\usepackage[hyphens]{url}  
\usepackage{graphicx} 
\usepackage{natbib}  
\usepackage{caption} 
\usepackage{xcolor}
\usepackage{algorithm}
\usepackage{algorithmic}
\usepackage{cite}
\usepackage{amsmath}
\usepackage{amsthm}
\usepackage{amssymb}
\usepackage{graphicx}
\usepackage{multirow}
\usepackage{booktabs}
\usepackage{svg}
\usepackage{pifont}

\usepackage{newfloat}
\usepackage{listings}

\DeclareCaptionStyle{ruled}{labelfont=normalfont,labelsep=colon,strut=off} 
\floatstyle{ruled}
\newfloat{listing}{tb}{lst}{}
\floatname{listing}{Listing}
\title{LAKD-Activation Mapping Distillation Based on Local Learning}
\author{
    \\
    Yaoze Zhang\textsuperscript{1}\equalcontrib,
    Yuming Zhang\textsuperscript{2}\equalcontrib,
    Yu Zhao\textsuperscript{1},
    Yue Zhang\textsuperscript{1},
    Feiyu Zhu\textsuperscript{3}
    \thanks{ Corresponding\,author:\,Feiyu\,Zhu,\,email:\,zeurdfish@gmail.com}
}

\nocopyright
\affiliations{
    \textsuperscript{\rm 1}Shanghai University of Science and Technology\\
    \textsuperscript{\rm 2}Southeast University\\
    \textsuperscript{\rm 3}Attrsense\\

}

\usepackage{bibentry}

\begin{document}

\maketitle

\begin{abstract}
Knowledge distillation is widely applied in various fundamental vision models to enhance the performance of compact models. Existing knowledge distillation methods focus on designing different distillation targets to acquire knowledge from teacher models. However, these methods often overlook the efficient utilization of distilled information, crudely coupling different types of information, making it difficult to explain how the knowledge from the teacher network aids the student network in learning. This paper proposes a novel knowledge distillation framework, Local Attention Knowledge Distillation (LAKD), which more efficiently utilizes the distilled information from teacher networks, achieving higher interpretability and competitive performance. The framework establishes an independent interactive training mechanism through a separation-decoupling mechanism and non-directional activation mapping. LAKD decouples the teacher's features and facilitates progressive interaction training from simple to complex. Specifically, the student network is divided into local modules with independent gradients to decouple the knowledge transferred from the teacher. The non-directional activation mapping helps the student network integrate knowledge from different local modules by learning coarse-grained feature knowledge. We conducted experiments on the CIFAR-10, CIFAR-100, and ImageNet datasets, and the results show that our LAKD method significantly outperforms existing methods, consistently achieving state-of-the-art performance across different datasets.

\end{abstract}

\section{Introduction}

Over the past decade, Knowledge Distillation (KD) has achieved significant advancements across various domains. It is widely applied in diverse domains, including image classification \cite{selfk,hint}, object detection \cite{fine-grained,localization}, and the compression of pre-trained models \cite{distilbert}.

The KD methods can be categorized into feature distillation, logit distillation, and attention distillation. Fundamentally, these approaches aim to bridge the gap between the student model and the teacher model by designing various distillation objectives, with the expectation of transferring the ``dark knowledge'' from the large teacher model. By tightly coupling multiple types of information through multiple targets \cite{selfkd,rkd}, these methods seek to enhance distillation performance.
\begin{figure}[t]
    \centering
    \includegraphics[width=1\linewidth]{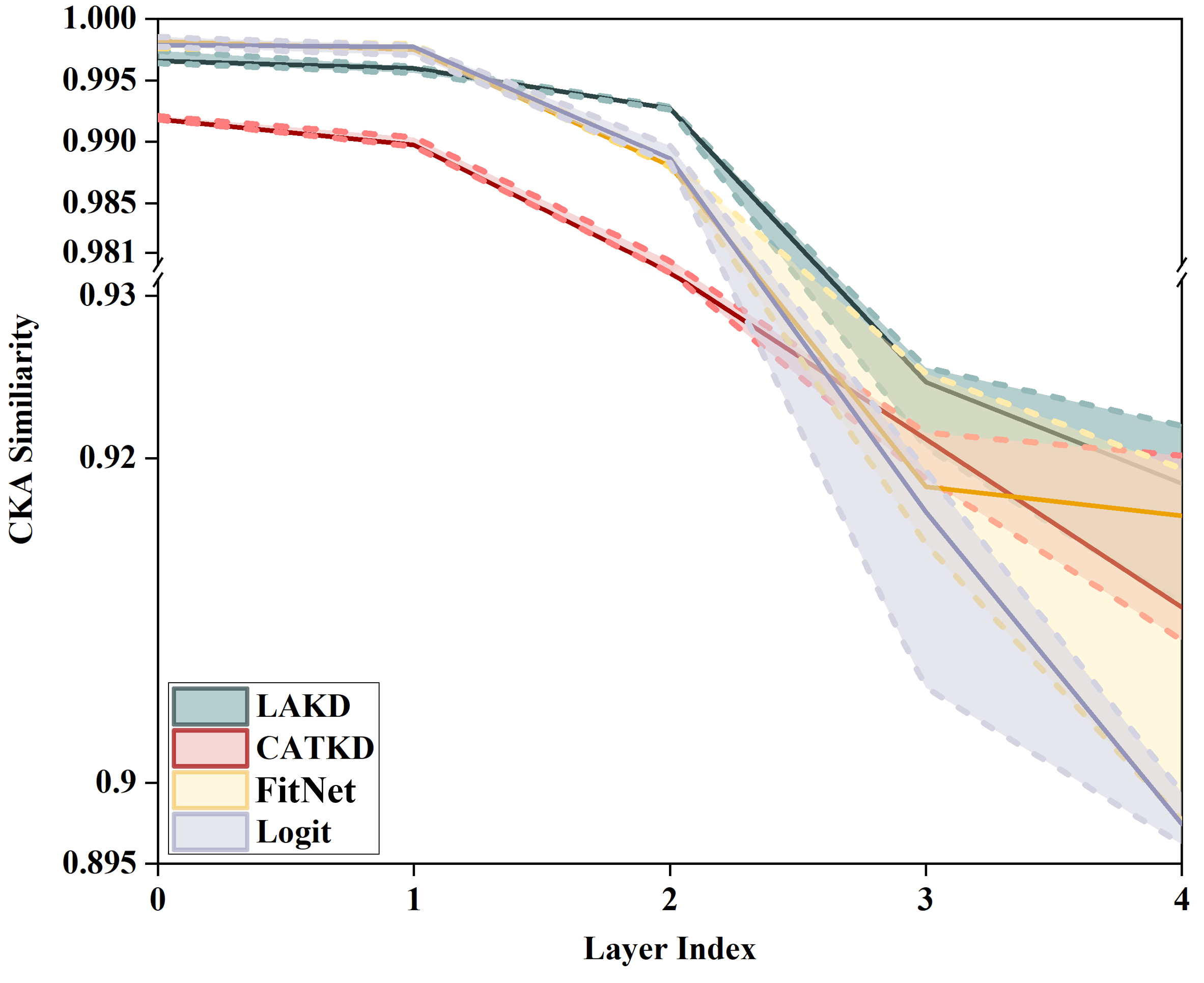}
    \caption{Comparison of teacher-student Layer Centered Kernel Alignment (CKA) \cite{kornblith2019similarity} similarity across different distillation methods. The teacher model is ResNet56, and the student model is ResNet20, using the following distillation methods: LAKD (Ours), CATKD \cite{catkd}, FitNet \cite{hint}, and KD \cite{hint}. FitNet aligns each layer individually.}
    \label{fig:cka}
\end{figure}

Although KD has achieved considerable success, previous methods are based on an implicit assumption: the learning objectives of the entire network are essentially independent and do not affect each other. Such assumptions can lead to issues where shallow features influence deeper structures during backpropagation, and the objectives of shallow layers are inconsistent with those of deeper layers \cite{ionescu2015matrix,hu2021hierarchical}. In the self-distillation process of existing methods, \cite{selfkd} introduces a branch at each stage, aligning each branch's feature maps and logits with those of the teacher model. Similarly, in OFD \cite{ofd} and simKD \cite{simKD}, both shallow and deep objectives involve aligning the feature maps of the current layer with those of the teacher model. These methods, due to the coupling of learning objectives during network propagation, hinder the transmission of information during distillation, negatively impacting model performance.

This paper introduces a novel framework called Local Attention Knowledge Distillation (LAKD), designed to enhance distillation interpretability and performance by leveraging attention maps and decoupling model components. LAKD consists of two key modules: Separation-Decoupling Mechanism and Non-Directional Activation Mapping.

The Separation-Decoupling Mechanism addresses coupling issues by applying gradient truncation, which divides the student model into independent modules that focus on learning layer-specific knowledge, reducing interference from shallow features, and improving representation integrity. However, simply decoupling and training these modules in isolation may lead to overfitting the teacher's features, reducing the student model's generalization ability. To overcome this, Non-Directional Activation Mapping uses the teacher model's attention to guide the student model, ensuring it focuses on critical features and refines its learning. The combination of these mechanisms achieves effective model decoupling while enhancing the student's learning process. As shown in Figure \ref{fig:cka}, LAKD exhibits higher CKA similarity with the teacher model, leading to improved performance compared to traditional feature distillation methods.

Our main contributions are summarized as follows: 

\begin{itemize}
\item We propose the LAKD framework, which more effectively utilizes the knowledge distilled from the teacher network and further enhances the interpretability during the distillation process.
\item LAKD introduces a separation-decoupling mechanism and non-directional activation mapping that can decouple and utilize distillation information, enabling the student model to focus on specific information.
\item Based on extensive experiments, our framework significantly outperforms existing methods on CIFAR-100, and ImageNet, consistently achieving state-of-the-art performance across different datasets.
\end{itemize}

\section{Related Work}
\subsection{Knowledge Distillation}

Knowledge distillation is first proposed by \cite{hinton}, leading to the development of logit distillation, feature distillation, and attention distillation. Among these, attention distillation gains particular attention due to its superior interpretability, guiding methods such as AT \cite{at} and CAT-KD \cite{catkd} to achieve state-of-the-art performance. However, with the advancement of knowledge distillation, the impact of coupling different objectives on model performance garners increasing attention. NKD \cite{nkd} decouples the target class information from the non-target class information and uses the decoupled information to guide the model. MasKD \cite{maskd} separates different pixels within a region for object detection tasks and employs masked tokens based on attention to assist in distillation. SDD \cite{sdd} decouples logits at the spatial level. Existing decoupling methods typically perform task decoupling at the image input or logit output stages, but information coupling still occurs during feature transmission. Unlike these current methods, we propose a Separation-Decoupling Mechanism that uses a local learning approach, enabling different layers to focus solely on their specific objectives, thereby achieving decoupling at the model level.

\subsection{Local Learning}
Local learning, as an innovative deep learning algorithm, emerges as a more memory-efficient and biologically plausible alternative to the end-to-end (E2E) training paradigm. However, the performance and applicability of local learning are limited by the design of auxiliary networks. Recently, methods like HPFF \cite{su2024hpff}, MAN \cite{su2024momentum}, and MLAAN \cite{zhang2024mlaan} achieve performance comparable to backpropagation (BP) by decoupling information. GLCAN \cite{zhu2024glcan} extends local learning to all image tasks, but the slowdown caused by auxiliary networks remains unresolved. In this paper, we introduce local learning into knowledge distillation (KD) tasks, decoupling complex teacher features at the network level. This approach reduces memory usage and improves the performance of distillation networks without compromising speed.

\section{Method}

\subsection{Preliminaries}
To clarify the background, in this section, we briefly outline the traditional distillation framework and explain its objectives and iterative process. We represent a data sample as \( x \) with its corresponding true label \( y \), and we use \( \mathcal L(y, y_s) \) to evaluate the performance of the entire network. In the distillation task, we denote the student model as \( S_{net} \) and the teacher model as \( T_{net} \), where \( T_{net} \) is considered a network with superior performance under the \( \mathcal L \) evaluation, while \( S_{net} \) represents a network with relatively weaker performance under the \( \mathcal L \) evaluation. The distillation network is divided into two parts: \( T_{net} \) is responsible for feature extraction, which can be expressed as \( F\{y_1, y_2, \dots, y_n\} = T(x) \), where \( F \) represents the set of learnable features generated by the teacher network, and \( T(.) \) denotes the forward computation of \( T_{net} \). \( S_{net} \) is responsible for learning the teacher's feature set \( F \). We denote the forward computation of the \( i \)-th layer of the student and teacher as \( F_S^l(i) \) and \( F_T^l(i) \) respectively, with the teacher's output denoted as \( y_t \). Finally, the performance of the entire student network is evaluated using \( \mathcal L(y, y_s) \). 

The premise of distillation lies in the fact that the information inherent in true labels is often challenging for lower-performing students to assimilate. Conversely, high-performing networks possess a wealth of learnable information beyond mere categorical data. Consequently, the student's objective can be approximated as improving their performance by imitating the teacher, as shown:
\begin{equation}
    \mathcal L(y, y_s) \approx \mathcal L(y_t, y_s)
\end{equation}

In conventional knowledge distillation, the approach typically involves minimizing \( \mathcal L(y_t, y_s) \) by reducing the inter-layer distance between the student and teacher networks. Specifically, this is achieved by minimizing the distance between the feature maps of corresponding layers in the student and teacher networks. The loss function is formulated as follows:
\begin{equation}
    \mathcal L(y_t, y_s)=\sum_{l \in \mathcal{L}} \frac{1}{N_l} \sum_{i=1}^{N_l} \left\| y_i - F_S^l(i) \right\|_2^2
\end{equation}

The term \(\sum_{l \in \mathcal{L}} \frac{1}{N_l} \sum_{i=1}^{N_l} \left\| y_i - F_S^l(i) \right\|_2^2\) represents the total feature-based loss across all layers.

In traditional knowledge distillation, the focus is primarily on minimizing a total loss that aggregates the contributions from multiple layers and outputs. This total loss\(\mathcal L_{\text{t}}\) is typically composed of the following components:
\begin{equation}
  \mathcal L_{\text{t}} = \alpha \mathcal L_{\text{h}} + (1 - \alpha) \mathcal L_{\text{s}} + \beta \sum_{l \in \mathcal{L}} \frac{1}{N_l} \sum_{i=1}^{N_l} \left\| y_i - F_S^l(i) \right\|_2^2  
\end{equation}
Where \( \mathcal L_{\text{h}} \) is the cross-entropy loss between the student model's predictions and the true labels. \( \mathcal L_{\text{s}} \) is the Kullback-Leibler divergence between the softmax outputs of the teacher and student models. \( \alpha \) and \( \beta \) are hyperparameters that control the weighting of the hard loss, soft loss, and feature-based loss, respectively.

In this approach, the backpropagation of the loss through the network computes the gradients for all layers based on this aggregated loss function:
\begin{equation}
    \theta_S \leftarrow \theta_S - \eta \frac{\partial \mathcal L_{\text{t}}}{\partial \theta_S} \label{eq:df}
\end{equation}

where \( \theta_S \) represents the parameters of the student model, and \( \eta \) is the learning rate.

\subsection{Local Attention Knowledge Distillation}
In existing distillation architectures, due to the high coupling of different losses during backpropagation, as shown in Equation \ref{eq:df}, the model's parameter updates are influenced by the interactions between different loss characteristics, ultimately leading to suboptimal learning performance.

\begin{figure*}[!h]
    \centering
    \includegraphics[width=1\linewidth]{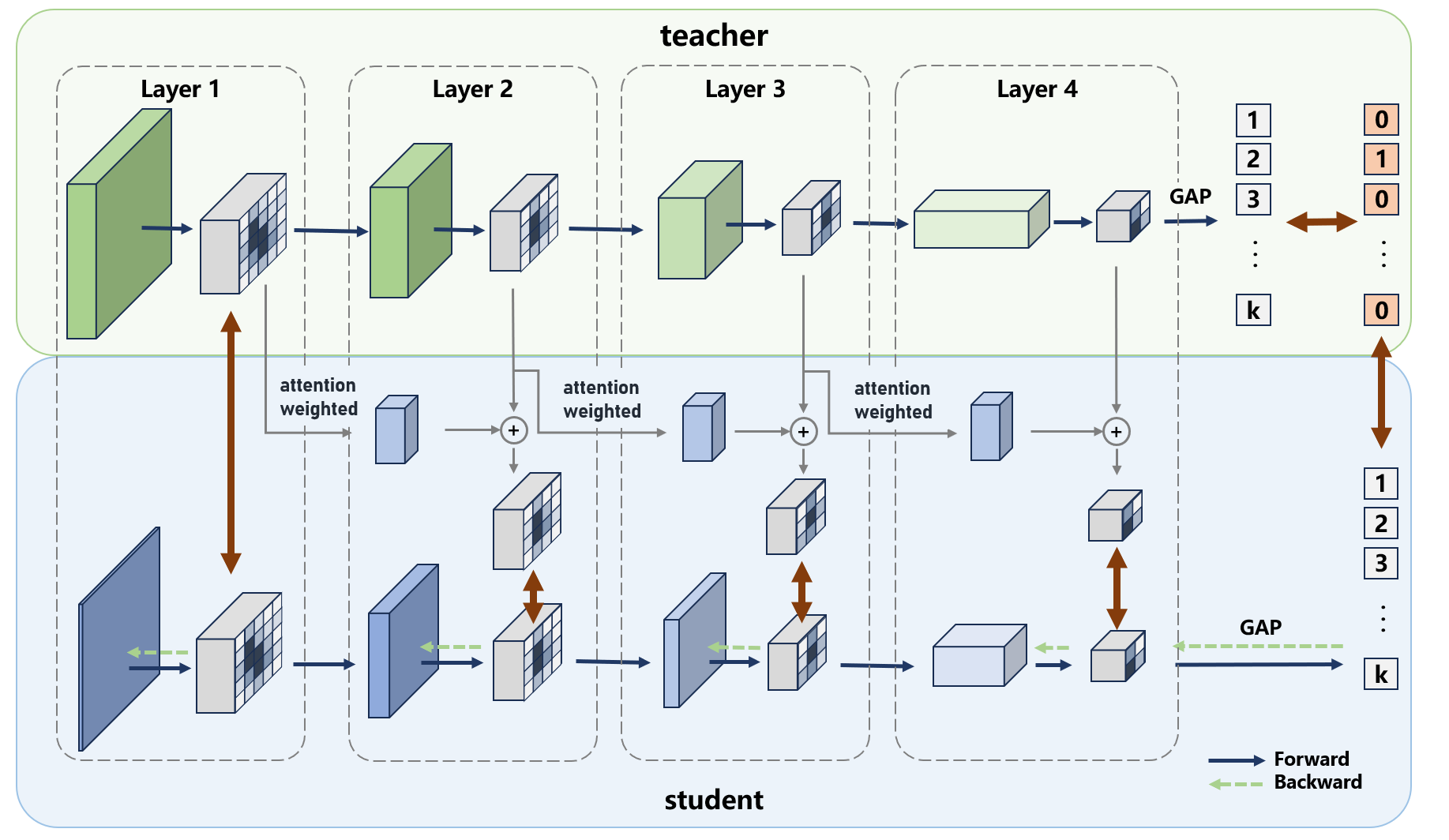}
    \caption{Overview of the proposed LAKD framework. Our method combines the Separation-Decoupling Mechanism (SDM) and Non-Directional Activation Mapping (NDAM). Building on CAT-KD, we add feature distillation for earlier layers. SDM applies gradient detachment to isolate the alignment tasks for each layer, while NDAM uses weights to integrate prior information and guide each module to focus on critical regions identified by the teacher.}
    \label{fig:lakd}
\end{figure*}

To address this, we propose a decoupled distillation framework called Local Attention Knowledge Distillation (LAKD), as depicted in Figure \ref{fig:lakd}. In LAKD, we utilize a Separation-Decoupling Mechanism (SDM) to align the features between the teacher and student models and apply Non-Directional Activation Mapping (NDAM) to optimize attention across the entire network.

\subsubsection{Separation-Decoupling Mechanism}
SDM is a greedy local attention knowledge distillation method introduced in our approach works by dividing the entire network into gradient-independent local networks, where feature-based losses are calculated and minimized independently in each local network. This method differs from traditional methods in that it does not aggregate cross-layer losses into a single objective function. Instead, the model is trained layer by layer, allowing for a more focused and potentially more efficient alignment of features at each stage.

The layer-wise feature loss \( \mathcal L_{\text{f}}^l \) for a specific layer \( l \) is defined as:
\begin{equation}
    \mathcal L_{\text{f}}^l = \frac{1}{N_l} \sum_{i=1}^{N_l} \left\| F_T^l(i) - F_S^l(i) \right\|_2^2
\end{equation}

Simultaneously, through the introduction of attention loss\( \mathcal L_{\text{att}} \) \cite{catkd}. The total loss of LAKD can be expressed as:
\begin{equation}
    \mathcal L_{\text{t}} = \alpha \mathcal L_{\text{h}} + (1 - \alpha) \mathcal L_{\text{att}} + \beta \mathcal L_{\text{f}}^l
\end{equation}

\begin{figure}[!h]
    \centering
    \includegraphics[width=1\linewidth]{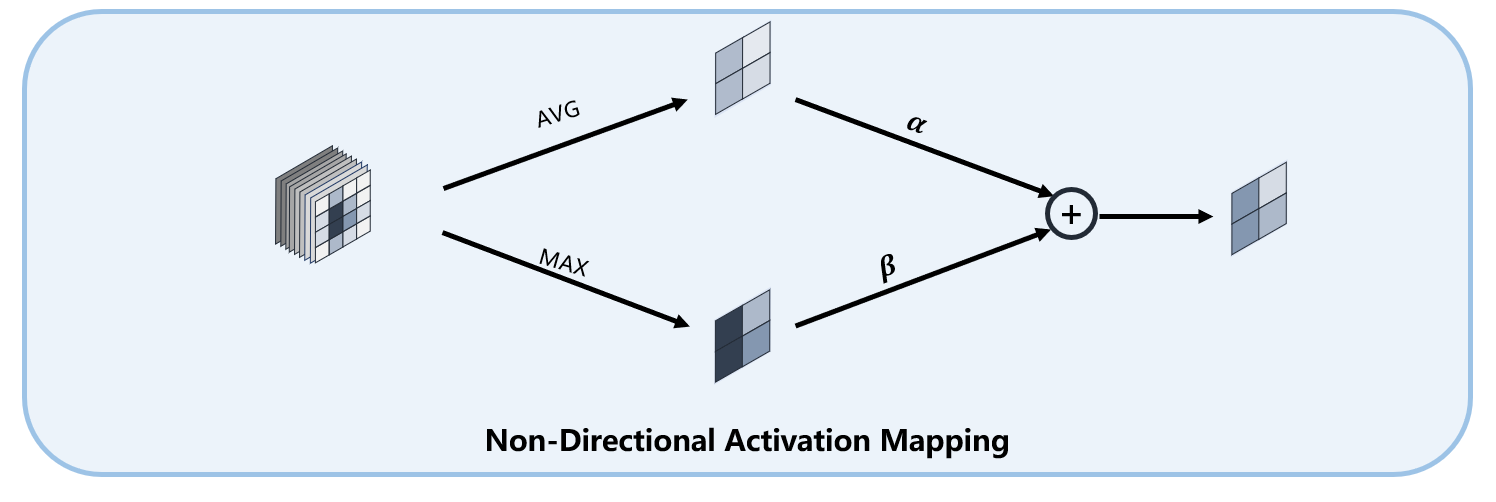}
    \caption{Illustration of the Non-Directional Activation Mapping (NDAM) Module. The module integrates feature maps using both average pooling (AVG) and maximum pooling (MAX) operations. The outputs are then combined to generate a refined activation map that highlights key regions of the input feature map.}
    \label{fig:ndam}
\end{figure}

In SDM, instead of summing these losses across all layers, the student model's parameters \( \theta_S^l \) are updated for each layer \( l \) individually based on the gradient of \( \mathcal L_{\text{f}}^l \):
\begin{equation}
    \theta_S^l \leftarrow \theta_S^l - \eta \frac{\partial \mathcal L_{\text{f}}^l}{\partial \theta_S^l}
    \end{equation}

In this layer-wise approach, the backpropagation algorithm is applied independently at each layer:

\begin{equation}
   \delta^l = \frac{\partial \mathcal L_{\text{f}}^l}{\partial F_S^l} 
\end{equation}

\begin{equation}
    \frac{\partial \mathcal L_{\text{f}}^l}{\partial \theta_S^l} = \delta^l \cdot \frac{\partial F_S^l}{\partial \theta_S^l}
\end{equation}

Here, the gradient \( \frac{\partial \mathcal L_{\text{f}}^l}{\partial \theta_S^l} \) is computed separately for each layer, allowing the model to adjust its parameters in a layer-specific manner.

SDM lies in its ability to focus on individual layers during the training process. By isolating the feature alignment loss at each layer and adjusting the student model accordingly, this approach allows for a more granular and more effective transfer of knowledge.

\subsubsection{Non-Directional Activation Mapping}
However, due to the excessive alignment between the student and teacher models' outputs when using SDM, the student model becomes overly dependent on certain features of the teacher model, lacking the ability to learn knowledge that the teacher model does not possess. This ultimately results in a decline in the student model's performance in real-world applications. We refer to this knowledge that the teacher model lacks as Extracurricular Knowledge(EK). The definition of EK is as follows:
\begin{equation}
   EK=\frac{\sum_{i=1}^{n}  {|| O^T_i \neq t_i\ and \ O^S_i == t_i ||}  }{|| O^T_i \neq t_i\ ||}
\end{equation}

Where \( n \) represents the total number of samples, \( O^T_i \) and \( O^S_i \) denote the predictions of the teacher and student models, respectively, for the \( i \)-th sample, and \( t_i \) represents the true label of the \( i \)-th sample. 

To address the aforementioned issue, we propose the NDAM method, as illustrated in Figure \ref{fig:ndam}. NDAM incorporates an attention mechanism based on \( L_{\text{att}} \). Leveraging the spatial invariance of convolution, we use the previously introduced attention information to weight the current feature map, thereby reducing the overfitting of the student model to the teacher model. In subsequent experiments, we validated the effectiveness of this method. The implementation of NDAM is as follows:
\begin{equation}
    F_{sum}(T)=\sum_{i=1}^{C}|T_i|
\end{equation}
\begin{equation}
    W=\alpha \times avgpool(F_{sum}(T)) + \beta \times maxpool(F_{sum}(T))
\end{equation}

Where \( T \) belongs to \( \mathbb{R}^{C \times H \times W} \), \( T_i = T(i,:,:)\). The operations \( \text{avgpool} \) and \( \text{maxpool} \) correspond to average pooling and max pooling, respectively. \( \alpha \) and \( \beta \) are coefficients that adjust the contribution of each block.

Essentially, the proposed SDM method effectively addresses the inherent coupling issues in traditional distillation frameworks by decoupling the various optimization objectives. At the same time, it introduces NDAM to mitigate the overfitting problem caused by different local blocks, ensuring that the output of the entire student network is more aligned with the global objective rather than the teacher network.

In addition to the two innovative methods mentioned above, LKAD has demonstrated outstanding performance across various datasets, showcasing its powerful capabilities.

\begin{table*}[htbp]
    \centering
    \setlength{\tabcolsep}{1.5mm}
    \renewcommand{\arraystretch}{1.2}
    \begin{tabular}{cccccccc}
    \hline
     & Teacher & ResNet56 & ResNet110 & ResNet32×4 & WRN-40-2 & WRN-40-2 & VGG13 \\
    
    Distillation & Acc & 72.34 & 74.31 & 79.42 & 75.61 & 76.61 & 74.64 \\
    \cmidrule(lr){2-8}
    Mechanism& Student & ResNet20 & ResNet32 & ResNet8×4 & WRN-16-2 & WRN-40-1 & VGG8 \\
    & Acc & 69.06 & 71.14 & 72.50 & 73.26 & 71.98 & 70.36 \\
    \hline
    \multirow{1}{*}{Logits} & KD \cite{hinton} & 70.66 & 73.08 & 73.33 & 74.92 & 73.54 & 72.98 \\
    \hline
    \multirow{5}{*}{Feature} & CRD \cite{crd} & 71.16 & 73.48 & 75.51 & 75.48 & 74.14 & 73.94 \\
    & OFD \cite{ofd} & 70.98 & 73.23 & 74.95 & 75.24 & 74.33 & 73.95 \\
    & FitNet \cite{hint} & 69.21 & 71.06 & 73.50 & 73.58 & 72.24 & 71.02 \\
    & RKD \cite{rkd} & 69.61 & 71.82 & 71.90 & 73.35 & 72.22 & 71.48 \\
    \hline
    \multirow{2}{*}{Attention} & AT \cite{at} & 70.55 & 72.31 & 73.44 & 74.08 & 72.77 & 71.43 \\
    & CAT-KD \cite{catkd} & 71.62 & 73.62 & \textbf{76.91} & 75.60 & \textbf{74.82} & \textbf{74.65} \\
    \hline
    \multirow{2}{*}{Local}& LAKD  & \textbf{72.01} & \textbf{74.26} & 75.72 & \textbf{75.87} & 74.47 & 74.45 \\
    & ↑ & +2.95 & +3.12 & +3.22 & +2.61 & +2.49 & +4.09 \\
    \hline
    \end{tabular}
    \caption{Results on CIFAR-100. Teachers and students have the same architecture. The best is marked in \textbf{bold}.↑ represents the performance improvement of LAKD compared with Student.}
    \label{tab:cifar100}
\end{table*}

\begin{table*}[htbp]
    \centering
    \setlength{\tabcolsep}{2mm}
    \renewcommand{\arraystretch}{1.1}
\begin{tabular}{cc|ccccccc}
\hline
& Teacher & Student & OFD & CRD  & KD  & AT & CAT-KD & LAKD \\ \hline
Top-1 & 73.31 & 69.75 & 70.81 & 71.17  & 70.66 & 70.69 & 71.26 & \textbf{71.51} \\ \hline
Top-5 & 91.41 & 89.07 & 89.98 & 90.13  & 89.88 & 90.01 & 90.45 & \textbf{90.46} \\ \hline
\end{tabular}
\caption{Comparison of Top-1 and Top-5 accuracy across various methods on Imagenet. In this group, we set ResNet34 as the teacher and ResNet18 as the student.}
\label{tab:imagenet}
\end{table*}

\section{Experiments}
In this section, we conduct experiments using three widely adopted datasets to verify the performance of our distillation framework: CIFAR-10, CIFAR-100 \cite{cifar}, and ImageNet \cite{im}. We use NVIDIA RTX 8000(46GB)GPU and test various network architectures, including ResNets \cite{resnet} of different depths, WRN \cite{wrn}, and VGG \cite{vgg}. We compare our method with several state-of-the-art distillation techniques based on different objectives, such as CRD \cite{crd}, OFD \cite{ofd}, FitNet \cite{hint}, RKD \cite{rkd},  AT \cite{at}, and CAT-KD \cite{catkd}. In our implementation, we employ the detach method and do not introduce local modules commonly used in local learning.

\subsection{Implement Details}

In our experiments on CIFAR-10 and CIFAR-100 datasets with ResNet-20, ResNet-32, ResNet8x4, ResNet-110, WRN-16-2, WRN-40-1 and VGG-8, we utilize the SGD optimizer with Nesterov momentum set at 0.9 and an L2 weight decay factor of 5e-4. We employ batch sizes of 64 for CIFAR-10 and CIFAR-100. The training duration spans 300 epochs, starting with initial learning rates of 0.05, following a linear decay learning rate strategy. For the student network, when using our method, decoupling makes the shallow layers easier to align. Therefore, we reallocate computational resources from the shallow layers to the deeper layers, allowing the aligned layers to shift forward during the distillation process. For example, while the basic feature alignment is performed at the n-th, 2n-th, and 3n-th layers, our method aligns at the 1st, (n+1)-th, and 3n-th layers.

In our experiments on Imagenet1K datasets with ResNet-18, we utilize the SGD optimizer with Nesterov momentum set at 0.9 and an L2 weight decay factor of 1e-4. We employ batch sizes of 128 for Imagenet1K. The training duration spans 100 epochs, starting with initial learning rates of 0.1, following a linear decay learning rate strategy. When using our method, our alignment layers are set to the 1st, 2nd, 4th, and 8th layers.
\subsection{Contrastive Results}
\subsubsection{Results on CIFRA100}

We first evaluate the effectiveness of LAKD on CIFAR-100 \cite{cifar}. To ensure a fair comparison, all methods are tested under identical configuration parameters. The results are presented in Table \ref{tab:cifar100}.

It can be observed that LAKD achieves the best performance across the majority of architectures, with only slightly suboptimal results on ResNet32×4, WRN40, and VGG13. To further investigate the reasons behind this phenomenon, we conduct an in-depth analysis, the results of which are discussed in the subsequent experiment.

\begin{figure*}[htb]
    \centering
    \includegraphics[width=1\linewidth]{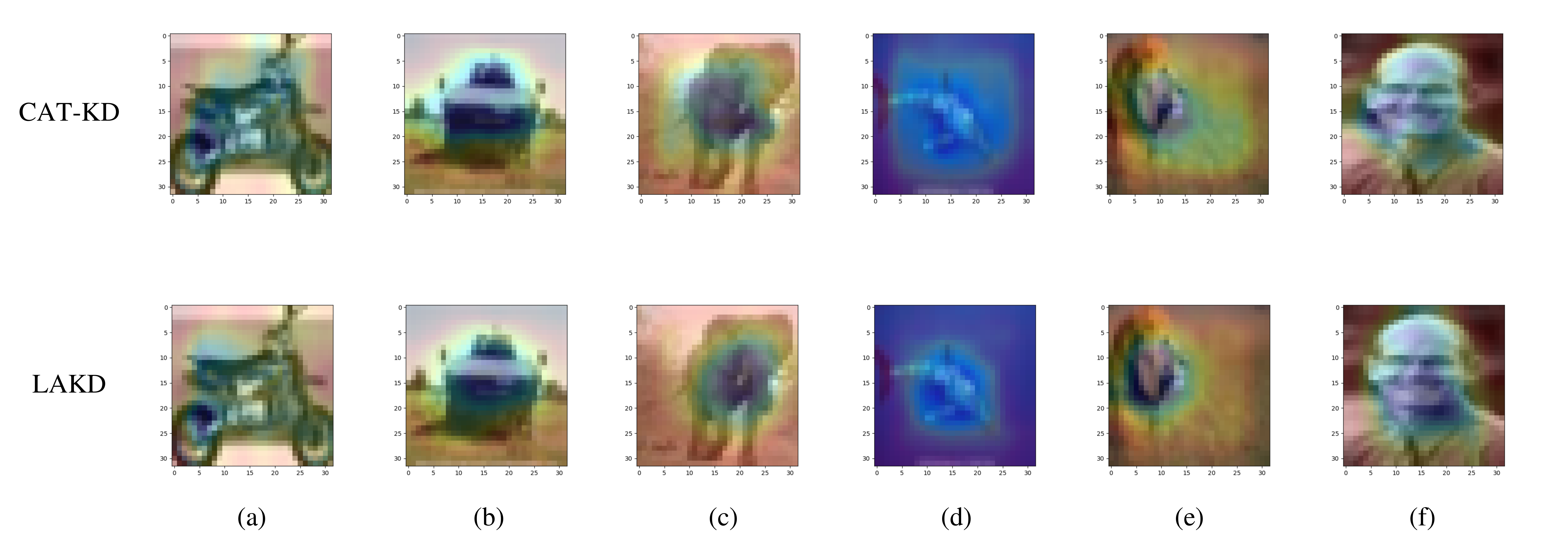}
    \caption{Top: Feature map of CAT-KD. Bottom: Feature map of LAKD.}
    \label{fig:fm1}
\end{figure*}

\subsubsection{Comparison of Different Detachment Module Settings}
\begin{table}[htb]
\centering
\begin{tabular}{cccc}
\toprule
Dataset & Detach & Location   & Top1 \\ \midrule
\multirow{4}{*}{\shortstack{\textbf{CIFAR-10}\\ Teacher (94.18\%)\\ CAT-KD  (93.2\%)}}  &        \ding{55} & [3,6,9]   & 92.88 \\ 
    &\ding{51} & [3,6,9]   & 93.08 \\ 
    &\ding{51} & [2,5,9]   & 93.21 \\ 
    &\ding{51} & [1,4,9]   & \textbf{93.27} \\ 
\bottomrule
\end{tabular}
\caption{Comparision of Detachment Module Setting}
\label{tab:a1}
\end{table}

\begin{figure}[htb]
    \centering
    \includegraphics[width=1\linewidth]{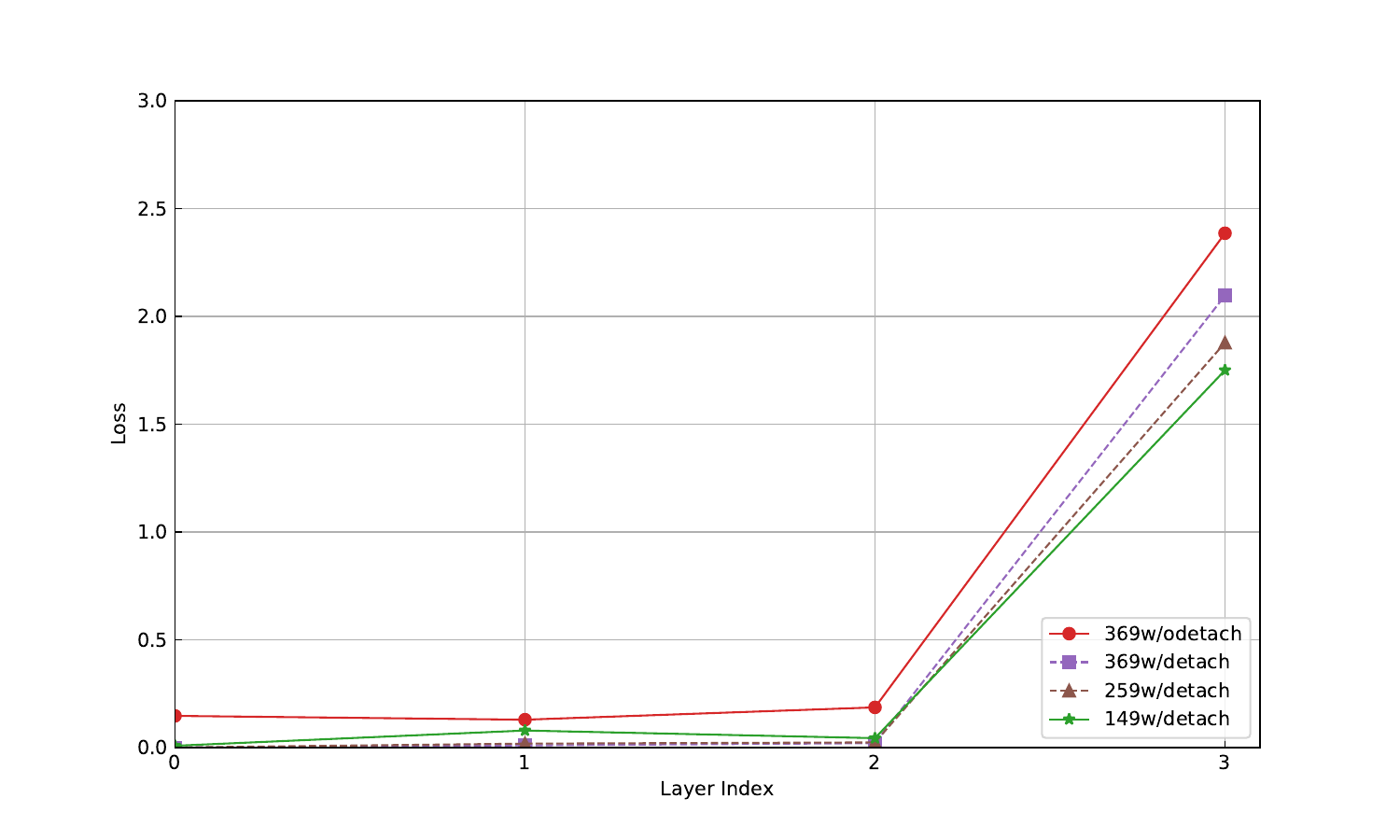}
    \caption{L2 loss of each layer of students and teachers under different Detachment Module settings}
    \label{fig:loss}
\end{figure}

We conduct experiments on CIFAR-10 using ResNet56 as the teacher and ResNet20 as the student to validate the effectiveness of our separation module. The results are shown in Table \ref{tab:a1}. Without gradient detachment, backpropagation with the loss from all layers yields a Top-1 accuracy of 92.88\%. After introducing the SDM, the Top-1 accuracy improves to 93.08\%. Furthermore, in our experiments, we find that further adjusting the gradient detachment points within the separation module and reallocating shallow computational resources to deeper layers could enhance performance. As shown in Figure \ref{fig:loss}, although reallocating shallow resources to deeper layers increases the challenge of shallow alignment, it effectively boosts deep alignment, leading to improved distillation outcomes. The experiments demonstrate that with appropriate SDM settings, LAKD achieves impressive results.

This phenomenon explains why LAKD exhibits suboptimal performance in certain experiments, as shown in Table 1. In our experiments, we ensure fairness by using consistent configurations across all methods. However, LAKD is particularly sensitive to the segmentation of different local modules. When tested with different models, mismatches in local module configurations may result in suboptimal performance.

\subsubsection{Comparison of Other Dataset}
We then evaluate our method on ImageNet, with comparison results presented in Table \ref{tab:imagenet}. Despite ResNet20 having fewer shallow resources available for deeper computations, our method still outperforms other KD methods.

\begin{figure*}[htb]
\begin{minipage}[b]{0.16\linewidth}
  \centering
  \includegraphics[width=2.5cm]{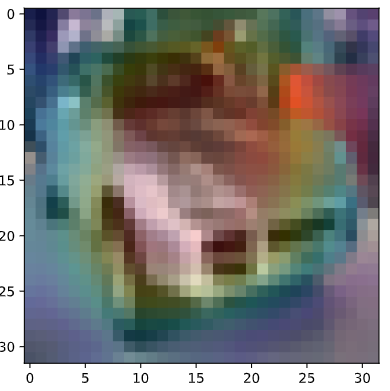}
  \centerline{(a)}\medskip
\end{minipage}
\begin{minipage}[b]{0.158\linewidth}
  \centering
  \includegraphics[width=2.5cm]{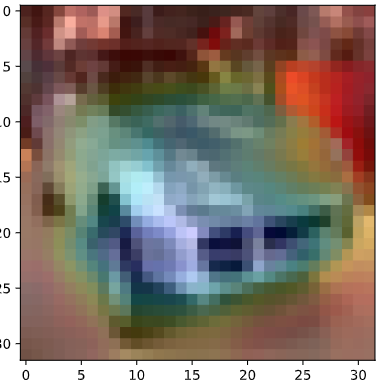}
  \centerline{(b)}\medskip
\end{minipage}
\begin{minipage}[b]{0.158\linewidth}
  \centering
  \includegraphics[width=2.5cm]{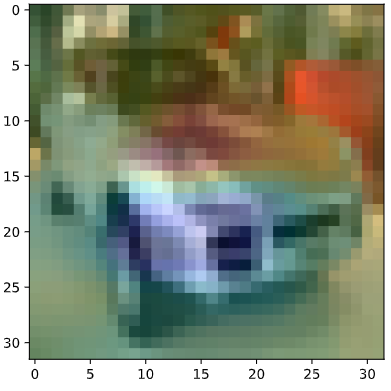}
  \centerline{(c)}\medskip
\end{minipage}
\begin{minipage}[b]{0.158\linewidth}
  \centering
  \includegraphics[width=2.5cm]{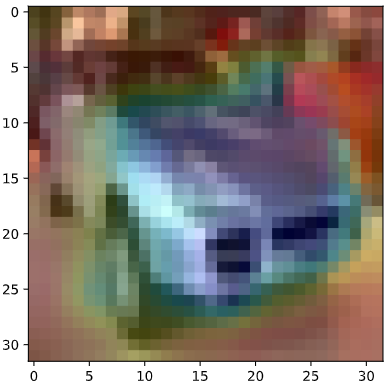}
  \centerline{(d)}\medskip
\end{minipage}
\begin{minipage}[b]{0.158\linewidth}
  \centering
  \includegraphics[width=2.5cm]{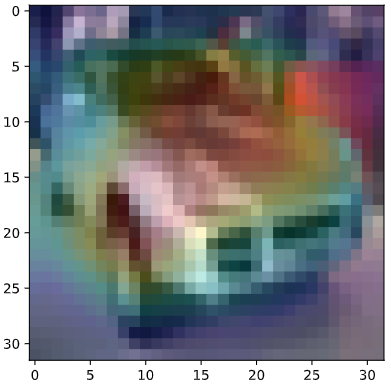}
  \centerline{(e)}\medskip
\end{minipage}
\begin{minipage}[b]{0.158\linewidth}
  \centering
  \includegraphics[width=2.5cm]{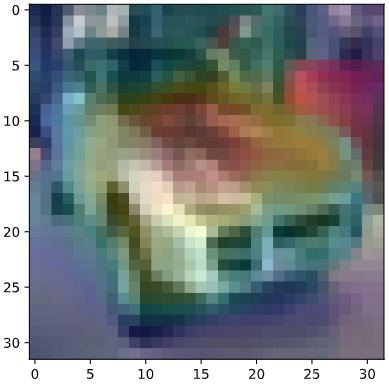}
  \centerline{(f)}\medskip
\end{minipage}
\caption{Visualization of feature maps. (a) Feature map of teacher. (b), (c), (d), (e), and (f) correspond to the settings in our ablation study on the non-directional activation configuration.}
\label{fig:fea}
\end{figure*}

\subsubsection{Memory Consumption}
We examine the GPU memory usage of Train From Scratch, CATKD, Normal Feature Distillation, and our method on CIFAR-100 and ImageNet, with results shown in Table \ref{tab:gpu_memory}. 
Using ResNet56 as the teacher and ResNet20 as the student, the previous distillation method increased memory usage compared to baseline training. In contrast, LAKD reduced memory usage by 17.1\%. Using ResNet34 as a teacher and ResNet18 as a student, similar results were observed on ImageNet, where LAKD reduced memory usage by 4.4\%. These findings show that although LAKD introduces more alignment information into the distillation process, it still exhibits better GPU memory efficiency than other methods. Since our approach reduces memory usage, we can use a more powerful but resource-intensive teacher for the same computational resources, which is also one of the advantages of LAKD.

\begin{table}[!h]
    \centering
      \setlength{\tabcolsep}{1mm} 
    \renewcommand{\arraystretch}{1.2}
    {
    \begin{tabular}{lll}
    \toprule
    \textbf{Dataset}  & \textbf{Method} & \textbf{GPU Memory(GB)} \\
    \midrule
    \multirow{4}{1.5cm}{CIFAR-10} 
         & Train From Scratch & 3.459G \\
         & CAT-KD & 3.610GG (↑4.3\%) \\
         & FitNet & 3.962G (↑14.5\%) \\
         & \textbf{LAKD} & \textbf{2.868G (↓17.1\%)} \\
    \midrule
    \multirow{4}{1.5cm}{ImageNet} 
         & Train From Scratch & 5.180G \\
         & CAT-KD & 5.544G (↑7.0\%) \\
         & FitNet & 6.544G (↑26.3\%) \\
         & \textbf{LAKD} & \textbf{4.950G (↓4.4\%)} \\
    \bottomrule
    \end{tabular}}
    \caption{Comparison of GPU Memory Usage between Different Methods. The batch size is set to 1024 on CIFAR-100 and 128 on ImageNet.}
    \label{tab:gpu_memory}
\end{table}

\subsection{Ablation Studies}
\subsubsection{Comparison of Features in Different Methods}
To demonstrate the advanced capabilities of our method, we compare the feature maps generated by our approach with those produced by the attention-based distillation method, CAT-KD. We use images from different categories in CIFAR-100. The resulting figures detailing these feature maps can be found in Figure \ref{fig:fm1}.

Upon analyzing them, we observe that in (a), (b), and (f), the feature distribution of LAKD over the target objects is more uniform compared to CAT-KD. In (c), (d), and (e), LAKD more accurately focuses on the critical regions of the target, with weaker attention given to irrelevant areas.

\subsubsection{Attention Effectiveness analysis}
\begin{figure}[htb]
    \centering
    \includegraphics[width=1\linewidth]{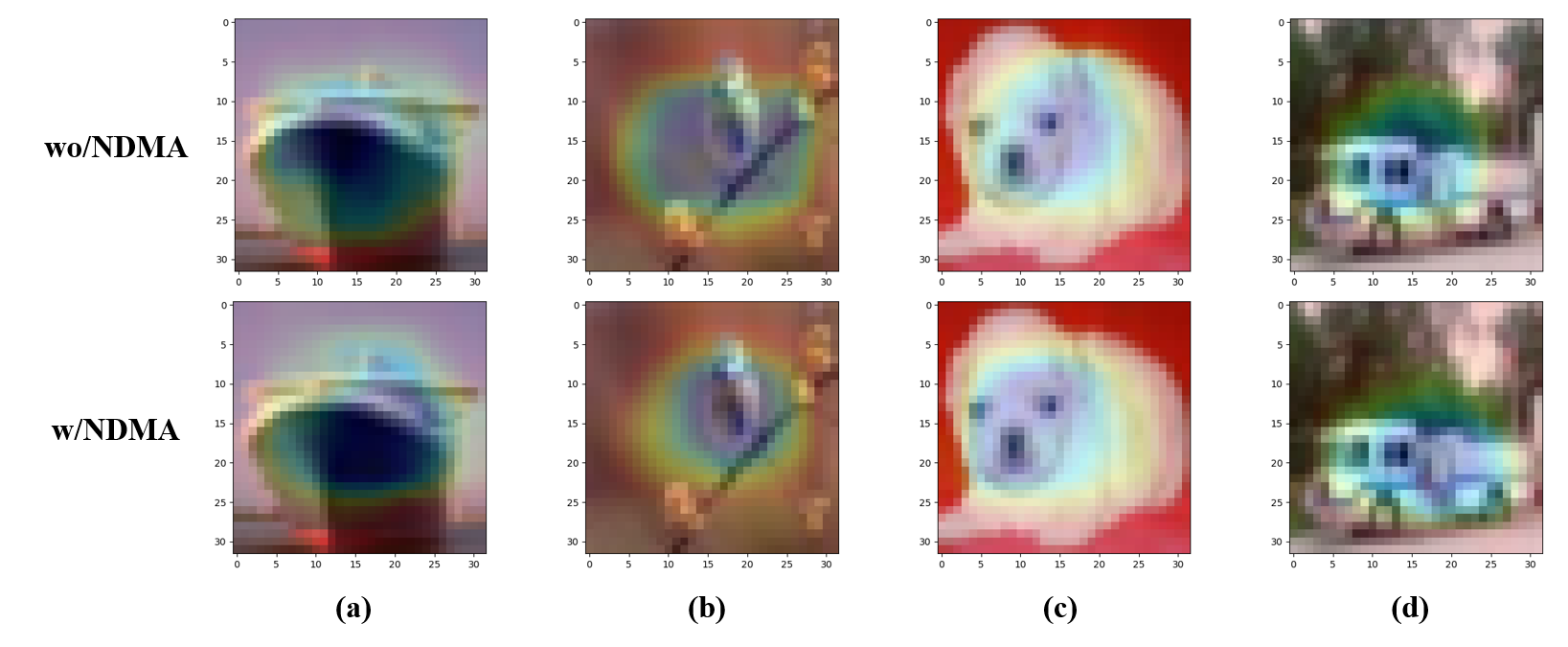}
    \caption{Top: Feature map of LAKD without NDAM. Bottom: Feature map of LAKD with NDAM.}
    \label{fig:fm2}
\end{figure}

To further validate the effectiveness of our method, we visualize the model's attention regions on CIFAR-10 using feature maps. We compare the results of the model with and without NDAM. Results are shown in Figure \ref{fig:fm2}.

NDAM significantly enhances the model's ability to identify critical regions. As seen in (a) and (b), with NDAM, the model's attention is more focused on the class-specific objects. Additionally, in (c) and (d), NDAM helps prevent the model's attention from drifting, thereby improving classification performance.

\subsubsection{Comparison of Weights in Attention Map}

We use a ResNet20 model with the Detachment Module applied at layers [3, 6, 9] as the baseline to test the performance of the Attention Map. The results are shown in Table \ref{tab:a2}. After incorporating the Non-Directional Activation, the model's performance improves significantly. The EK values indicate that taking the absolute value of feature map elements may introduce more Extracurricular Knowledge during the student's learning process from the teacher. Moreover, in line with the concept of attention maps, using max elements to represent focus areas further enhances the model's performance. Feature maps for the different methods are illustrated in Figure \ref{fig:fea}. Methods without Non-Directional Activation weighting are less effective in guiding the student model. Equal use of the maximum and average values in feature maps helps the student learn but may introduce irrelevant focus. Increasing the weight of the maximum value better aligns the student with the teacher.
\begin{table}[!h]
\centering
\begin{tabular}{cp{1.5cm}p{1.5cm}p{1cm}p{1cm}}
\toprule
abs & $\alpha$ & $\beta$  & EK(\%) & Top1 \\ \midrule
\ding{55} & 0 & 0  & 32.82 & 93.14 \\ 
\ding{51} & $0.5$ & $0.5$  & 31.61 & 93.22 \\ 
\ding{55} & $0.5$ & $0.5$  & 31.10 & 93.27 \\ 
\ding{51} & $0.25$ & $0.75$  & 35.74 & \textbf{93.39} \\ 
\ding{55} & $0.25$ & $0.75$  & 33.33 & 93.31 \\ \bottomrule
\end{tabular}
\caption{Ablation experiments on our non-directional activation setting. The abs represents whether to perform absolute value on the element }
\label{tab:a2}
\end{table} 

\section{Conclusion}
In this paper, we present Local Attention Knowledge Distillation (LAKD) for the first time, a novel framework applying local learning to knowledge distillation tasks. LAKD addresses performance issues in existing methods by reducing excessive information coupling. It features two components: the Separation-Decoupling Mechanism, which effectively mitigates the problem of excessive feature coupling in existing methods, and Non-Directional Activation Mapping, which combines previously introduced attention information to enhance information exchange between independent modules, guiding the student model to focus on critical regions. Tested on three major datasets, LAKD significantly improves the student model’s learning, surpassing current methods. After decoupling the modules in LAKD, attention maps facilitate information exchange between independent modules. However, attention maps may miss some teacher model details, potentially leading to a loss of distillation in formation. Future work will explore auxiliary networks to enhance information fusion during distillation.

\bibliography{ref}

\end{document}